\documentclass{article}
\usepackage[T1]{fontenc}
\usepackage{iclr2027_conference,times}
\usepackage{amsmath}
\usepackage{graphicx}
\usepackage{booktabs}
\usepackage{array}
\usepackage{longtable}
\usepackage{tabularx}
\usepackage{capt-of}
\usepackage{pifont}
\usepackage{hyperref}
\usepackage{url}
\usepackage{fvextra}
\usepackage{xcolor}
\newif\ifshowrev
\showrevfalse
\newcommand{\rev}[1]{\ifshowrev{\color{blue}#1}\else#1\fi}
\newenvironment{revblock}{\ifshowrev\color{blue}\fi}{}
\newcommand{\tabcite}[1]{~\citep{#1}}

\iclrfinalcopy

\begin{document}

\title{CompMat-Bench: Benchmarking AI Agents for Computational Materials Science}
\author{Chenmu Zhang, Levi Felix, Jun-Jie Zhang, Xingfu Li, Xuelian Jiang, \\
\bf Tao Jiang, Subhendu Mishra, Xixi Qin \& Boris Yakobson \\
Department of Materials Science and NanoEngineering \\
Rice University, Houston, TX 77005, USA \\
\texttt{zcmben2014@gmail.com, biy@rice.edu}
}
\maketitle
\lhead{Preprint}
\begin{abstract}
Evaluating AI agents on scientific research tasks is constrained by the time and
resources required for the underlying experiments or calculations.
In computational materials research, repeating the same expensive
simulations across agents and trials can make evaluation
impractical.
We introduce CompMat-Bench, a benchmark of 94 tasks derived from recently
published computational materials studies, each asking agents to complete a
step toward achieving the study's scientific goal.
We reproduce the research steps in advance and assess agents on preparing
inputs and analyzing outputs for expensive simulations, so expensive
simulations can be avoided during evaluation.
The reproduced inputs and results serve as ground truth for grading agents
with fixed rules, without an LLM judge.
\rev{We extend the benchmark to} four evaluation conditions: single tasks and
workflows composed of related tasks, each with full or reduced
methodological guidance.
\rev{In a workflow, the agent runs all simulations itself.}
With full guidance on single tasks, agents based on \rev{nine} LLMs demonstrate
the ability to complete individual materials research steps, with pass rates
of \rev{62.8--97.9\%} across 94 tasks.
Both longer workflows and reduced guidance can limit agent performance, but
in different ways for different agents: they lower the pass rates of the
weaker agents, whereas the \rev{stronger agents fall} only when a long workflow is
combined with reduced guidance.
Failure analysis attributes \rev{more} failures to scientific errors \rev{than}
to errors in software usage.
CompMat-Bench provides a basis for comparing agents on the steps of real
materials research and for analyzing agent failure modes.
\end{abstract}

\section{Introduction}
\label{sec:introduction}

Large language models (LLMs) have demonstrated strong coding capabilities,
including solving programming problems and fixing bugs in software
repositories
\citep{chen2021humaneval,jimenez2024swebench,yang2024sweagent,tian2024scicode}.
They have also demonstrated strong scientific understanding
\citep{rein2023gpqa,hendrycks2021mmlu}.
Computational materials research brings these capabilities together:
scientific understanding guides the choice of models and methods, while
code implements calculations and analyzes results.
Yet separate assessments of coding and scientific understanding do not
establish how well agents can combine them in materials research.
We therefore need benchmarks that directly quantify LLM agents' capabilities
in computational materials research and help assess their potential to
accelerate scientific research
\citep{lu2024aiscientist,boiko2023coscientist,ghafarollahi2025atomagents}.

\begin{figure}[t]
\centering
\includegraphics[width=\textwidth]{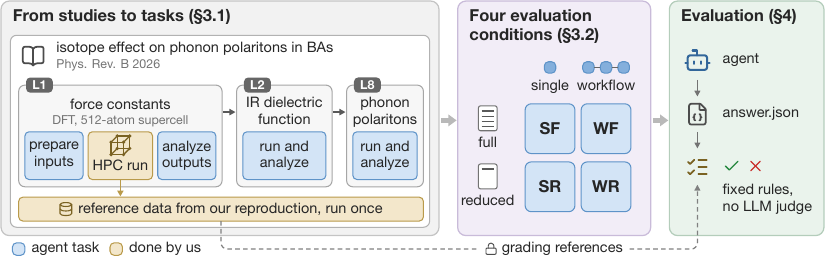}
\caption{Overview of CompMat-Bench, illustrated with research steps from
a study of isotope effects in boron arsenide \citep{wu2026isotope}.
For an expensive step, input preparation and output analysis are separate
tasks; prepared inputs are graded without running the simulation, and
analysis starts from our precomputed outputs.
Lighter calculations are executed by the agent.
Tasks are evaluated singly\rev{, or chained into a workflow in which the agent
runs all simulations itself,} with full or reduced method guidance.
Fixed rules grade submissions against hidden references from our
reproduction.}
\label{fig:benchmark-overview}
\end{figure}

However, evaluating agents on computational materials research is challenging
because simulations can take hours, or even days.
Repeating these simulations across agents and trials is costly and makes
rapid evaluation impractical \citep{cao2026simulcost}.
Most existing benchmarks either exclude computationally demanding tasks,
as in ScienceAgentBench \citep{chen2025scienceagentbench}, or provide
substantial computing resources (e.g., up to 72 hours per run in AutoMat
\citep{huang2026automat}).
The former narrows the range of research represented, while the latter
retains the cost of running simulations for each attempt.
Our aim is to evaluate LLM agents on \textbf{diverse research problems} in
computational materials research, \textbf{without expensive simulations}
during evaluation.

We introduce \textbf{CompMat-Bench}, a benchmark of 94 tasks derived from
15 recently published, peer-reviewed studies in computational materials research.
We remove expensive simulations from task evaluation by
(1) identifying the research steps that require them and reproducing
    these simulations in advance,
(2) saving their inputs and results as grading references, and
(3) assessing agents on preparing simulation inputs and analyzing the
    resulting data (Fig.~\ref{fig:benchmark-overview}).
Most computational materials tools are standard software packages with
fixed interfaces.
These interfaces, together with our reproduced inputs and results, allow
us to grade agents using fixed rules, including numerical tolerances for
scientific outputs, without an LLM judge.

CompMat-Bench also provides a protocol for controlling task difficulty
through methodological guidance and task span.
For each task, we provide full- and reduced-guidance prompts that preserve
the scientific target and grading criteria.
We combine scientifically dependent tasks from the same paper into eight
workflows, each requiring an agent to reach the final result from the
initial inputs\rev{, running all simulations itself}.
Crossing guidance with task span yields four evaluation conditions:
single tasks with full (SF) or reduced (SR) guidance, and workflows with
full (WF) or reduced (WR) guidance.

We evaluate \rev{ten} agents built on \rev{nine} LLMs \rev{(seven of them
under all four conditions)} and analyze their failures.
Our contributions are:

\begin{enumerate}
\renewcommand{\labelenumi}{(\theenumi)}
\item \textbf{Computational materials benchmark.} We introduce CompMat-Bench,
with 94 tasks for fixed-rule evaluation of computational materials research
without expensive simulations.

\item \textbf{Difficulty control.} We vary task difficulty along two axes,
methodological guidance and task span, while keeping the scientific target and
grading of each task fixed, which quantifies their individual contributions to
agent performance.

\item \textbf{Failure analysis.} We identify common failure modes of agents in
computational materials research: agents do not recognize anomalies in their
own results that an experienced researcher would notice, and agents based on
different LLMs lack the same implicit knowledge of software interfaces.

\end{enumerate}

\section{Related Work}
\label{sec:related-work}

\paragraph{Evaluation cost.}
Computational materials research relies on simulations, such as density
functional theory (DFT) and molecular dynamics (MD), that can run for hours
or even days on high-performance computing (HPC) clusters
\citep{camps2025quantum}.
A benchmark that includes these simulations must repeat them for every
agent, trial and evaluation condition.
Simulation time is an evaluation cost that token counts
\citep{kapoor2025agentsmatter} do not capture \citep{cao2026simulcost}.
Most existing benchmarks either avoid this cost or pay it in every attempt.
The first approach, \emph{exclusion}, keeps expensive computation out of
the tasks \citep{bogin2024super}.
FIRE-Bench selects low-cost machine-learning studies (most attempts finish
within one hour) \citep{wang2026firebench}.
CORE-Bench selects papers whose code runs in under 45 minutes
\citep{siegel2024corebench}.
ReplicationBench caps the core contributions of each paper at six hours
(with intermediate data products for costlier steps)
\citep{ye2025replicationbench}.
Exclusion keeps attempts cheap but limits the research to light computation.
The second approach, \emph{rerunning}, executes the computation in every
attempt \citep{chan2025mlebench,faroughy2026colliderbench}.
AutoMat allows each run up to 96 CPU cores and 72 hours
\citep{huang2026automat}.
PaperBench executes each submission on a fresh GPU machine
\citep{starace2025paperbench}.
Rerunning retains the research but repeats its cost for every agent and
trial.
Recent benchmarks take a third approach, \emph{precomputation}, and supply
outputs computed in advance.
gwBenchmarks builds its tasks on public gravitational-wave simulations and
catalogs \citep{islam2026gwbenchmarks}.
MDArena evaluates the stages before and after MD runs, which lack a ground
truth and are stochastic \citep{anand2026mdarena}.

CompMat-Bench runs each expensive simulation once in advance and evaluates
agents on preparing its inputs and analyzing its outputs.
This suits computational materials research because most simulations run in
standard software packages with fixed interfaces.
Input files that agents prepare can therefore be graded by fixed rules
without running them, and the outputs of one reproduction serve every agent
and trial.
Table~\ref{tab:prior-benchmarks} compares benchmarks on three properties:
tasks derived from published studies, tasks involving expensive computation,
and expensive computation excluded from evaluation.
Each prior benchmark lacks at least one property or holds it only in part,
whereas CompMat-Bench holds all three.

\begin{table}[t]
\centering
\caption{CompMat-Bench and prior agent benchmarks on three properties: tasks
derived from published studies, tasks involving expensive computation, and
expensive computation excluded from evaluation; the last column gives the
grading method.
\ding{51}~yes, \ding{55}~no, (\ding{51})~in part, --~no expensive
computation in the tasks.}
\label{tab:prior-benchmarks}
\small
\providecommand{\tabcite}[1]{}
\setlength{\tabcolsep}{3pt}
\begin{tabular}{@{}llcccl@{}}
\toprule
 & & Published & Expensive & Excluded from & \\
Benchmark & Domain & studies & computation & evaluation & Grading \\
\midrule
FIRE-Bench\tabcite{wang2026firebench} & ML & \ding{51} & \ding{55} & -- & LLM judge \\
ReplicationBench\tabcite{ye2025replicationbench} & astrophysics & \ding{51} & (\ding{51}) & (\ding{51}) & fixed rules \\
PaperBench\tabcite{starace2025paperbench} & ML & \ding{51} & (\ding{51}) & \ding{55} & LLM judge \\
AutoMat\tabcite{huang2026automat} & materials & \ding{51} & \ding{51} & (\ding{51}) & LLM judge \\
SimulCost\tabcite{cao2026simulcost} & physics sim. & \ding{55} & (\ding{51}) & \ding{55} & fixed rules \\
\midrule
\textbf{CompMat-Bench} & materials & \ding{51} & \ding{51} & \ding{51} & fixed rules \\
\bottomrule
\end{tabular}

\end{table}

\paragraph{Real research tasks.}
A benchmark based on real research tasks better reflects the research
capability of agents.
However, collecting and converting real research tasks is difficult, and
some benchmarks define their own tasks \citep{zhu2025critpt}.
SimulCost asks agents to tune the parameters of standard physics solvers
\citep{cao2026simulcost}.
gwBenchmarks defines eight gravitational-wave research tasks, from
regression to physics derivation \citep{islam2026gwbenchmarks}.
A better approach is to select published, peer-reviewed research and
convert it into benchmark tasks
\citep{qiu2026prbench,majumder2025discoverybench,deng2026qmpbench}.
ScienceAgentBench adapts 102 tasks from 44 peer-reviewed publications
\citep{chen2025scienceagentbench}.
CORE-Bench builds tasks from the released code of 90 papers
\citep{siegel2024corebench}.
PaperBench asks agents to replicate 20 ICML 2024 papers
\citep{starace2025paperbench}.
ReplicationBench selects 20 peer-reviewed astrophysics papers with their
authors' collaboration \citep{ye2025replicationbench}.
FIRE-Bench further decomposes each paper into a tree of research problems
\citep{wang2026firebench}.
CompMat-Bench converts peer-reviewed materials studies, mostly published
between January and March 2026, into research steps that keep their
scientific dependencies.
It reproduces the expensive simulations of these studies and uses the
results as ground truth, so that the ground truth is consistent with the
task settings and includes the details that papers omit.

\paragraph{Materials science benchmarks.}
Although the number of materials science benchmarks keeps growing
\citep{zaki2024mascqa,rubungo2025llm4matbench,yang2026autodft}, most of
them focus on a single type of computational approach or tool.
MatTools tests knowledge and use of the pymatgen library
\citep{liu2025mattools}.
INCARBench tests the preparation of VASP INCAR files against Materials
Project records \citep{shao2026incarbench}.
QMBench tests DFT knowledge, structure-file generation and parameter choice
without running calculations (free text graded by an LLM)
\citep{wang2025qmbench}.
DFTBench runs Quantum ESPRESSO workflows in every attempt
\citep{hu2026tritondft}.
MDGym reruns LAMMPS and GROMACS simulations in every attempt (for both
materials and biomolecular systems) \citep{kumar2026mdgym}.
AutoMat, the closest work, asks agents to reproduce 85 claims from 10
materials papers across several methods and scores them with an LLM judge
\citep{huang2026automat}.
Its claims are unevenly distributed across these papers, with at least 31
from a single paper (counted from the released dataset,
\citealp{huang2026automatdata}).
CompMat-Bench draws its 94 tasks from 15 published studies that use 12
scientific codes, with no study contributing more than 9 tasks, and grades
them with fixed rules instead of an LLM judge.

\section{Benchmark Construction}
\label{sec:construction}

Here, we describe how CompMat-Bench is constructed: how tasks are derived
from published studies (Section~\ref{sec:studies-to-tasks}), how guidance
and task span are controlled (Section~\ref{sec:guidance-span}), and what the
resulting tasks cover (Section~\ref{sec:composition}).

\subsection{From studies to tasks}
\label{sec:studies-to-tasks}

\paragraph{Study selection.}
We draw studies from four journals: npj Computational Materials, Physical
Review B, ACS Nano and Nano Letters.
We collected the 1,912 articles published from January to March 2026 whose
titles and abstracts were available through the OpenAlex API
\citep{priem2022openalex}.
We keep studies whose core is a computation on a real material with a
checkable result, and exclude experiment-led, dataset and review papers.
We then require a part of the study that we can afford to reproduce and
calculations that are light on the agent's side but demand scientific
reasoning.
We also exclude studies made public before August 2025 (e.g., a study
published in January 2026 whose preprint appeared in March 2025)
\citep{jain2025livecodebench} and studies that rely on closed-source codes.
These filters leave 72 studies, from which we select 14 by hand, favoring
methods and codes not yet covered.
A fifteenth study, published in September 2025, served as the pilot for
these criteria.

\paragraph{Decomposition and reproduction.}
We convert each study into tasks in two steps.
First, a language model guided by a structured prompt adapted from
FIRE-Bench \citep{wang2026firebench} decomposes the study into a directed
acyclic graph of research steps, which we check against the paper.
Each step is one calculation that produces a new physical quantity shown in
a main-text figure or table, and each edge records an upstream result that
the step uses.
These graphs have 6 to 18 steps per study.
Second, we select a logically connected sub-graph in which every selected
step keeps all of its upstream steps.
Each of the 15 studies contributes 3 to 7 steps, with a median of 5
(Table~\ref{tab:studies}).
Of all 65 steps, 30 need HPC resources for calculations such as DFT, and 35
run within minutes on one machine.
We reproduce these sub-graphs on HPC clusters and local workstations, and
our results serve as the grading references.
They sometimes differ from the published values, mostly because papers do
not report some inputs and we have to set them ourselves.

\begin{figure}[t]
\centering
\includegraphics[width=\textwidth]{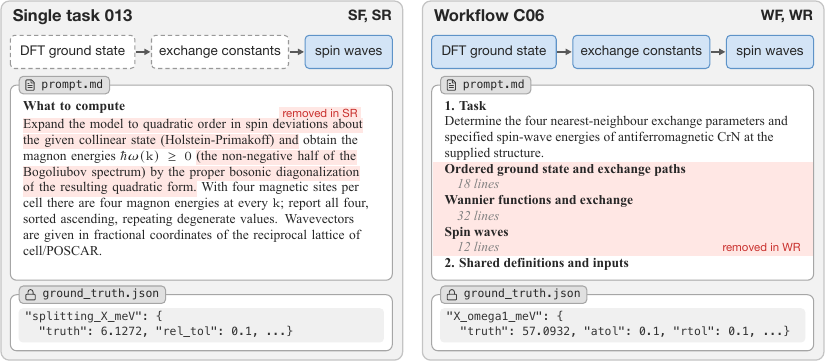}
\caption{A single task and a workflow from a study of magnons in
antiferromagnetic CrN \citep{kim2026magnons}.
Both panels show the same three stages.
Left: in single task 013 (SF, SR), the first two stages (dashed) are
precomputed, and the agent computes the spin waves from their outputs.
Right: in workflow C06 (WF, WR), the agent runs all three stages.
Shaded prompt text is removed in the reduced-guidance version, and the rest
of the prompt is identical.
Bottom: the hidden reference value and tolerance for one graded quantity.}
\label{fig:task-instance}
\end{figure}

\paragraph{Tasks and grading.}
We build three kinds of tasks.
For a step that needs HPC, the agent either prepares the simulation inputs
or analyzes our precomputed outputs (Fig.~\ref{fig:benchmark-overview}).
For a step that does not, the agent runs the calculation and analyzes the
result.
Fig.~\ref{fig:task-instance} shows a concrete example from a study of
magnons in CrN, published in Physical Review B \citep{kim2026magnons}.
Its three stages correspond to three tasks: preparing the DFT inputs,
deriving exchange constants from our precomputed DFT outputs, and computing
the spin-wave dispersion, including the magnon splitting at the zone
boundary.
The left panel of Fig.~\ref{fig:task-instance} shows the prompt of the last
task.
Its full-guidance prompt states the scientific target, the method and the
required output, and the reduced-guidance prompt omits how the magnon
energies are obtained, the Holstein-Primakoff expansion and the Bogoliubov
diagonalization (shaded).
The hidden reference stores our value for each graded quantity with its
tolerance, here the zone-boundary magnon splitting (6.1272~meV, within 10\%).
Input files are graded on the effective value of each required setting,
including software defaults, and on structures equivalent to the reference.
Analysis results must match our reference within numerical tolerances,
10\% relative by default.
A task scores 1 only when every check passes.
Tasks have 1 to 55 checks, with a median of 12.
\rev{Our own reproduction gives one complete set of inputs and its outputs.
This set is the reference for grading, but the grading rules are wider than
the reference.
The 94 tasks have 1,191 checks.
In general, they are of four kinds (Appendix~\ref{app:grading-checks}).
(1) Range: a check accepts an interval of values instead of one value.
For example, the result of a DFT calculation converges as the plane-wave
cutoff increases, so the rule sets a minimum for the cutoff.
For a quantity that the agent computes, the rule accepts a tolerance around
the value from our own reproduction, 10\% by default.
(2) Subset: only the settings that are decisive for the physical claim of the
original paper are checked.
For example, spin polarization is checked for a magnetic material and not for
a non-magnetic one.
(3) Physical equivalence: answers that are physically the same are accepted.
For example, a crystal structure shifted by a translation is the same
structure, although every atomic coordinate differs.
The grading rule here is a program that we wrote to test whether the
submitted structure is equivalent to our reference by a translation.
(4) Exact match: the answer must equal the reference.
This applies to answers with one correct value, such as integers, signs and
booleans.}
Prompts are written in our own words and structures are regenerated from
public databases, so no text, figures or values are copied from the source
papers.

\paragraph{Quality control.}
Each task is reviewed by a computational materials researcher with a PhD
in a related field, who grades it on four questions:
(1) answer leakage, whether any file given to the agent reveals the answer,
the method or a shortcut;
(2) prompt clarity, whether the prompt defines the task and its valid
answers without ambiguity;
(3) reference quality, whether the reference answer or input set is
physically reasonable; and
(4) scoring fairness, whether an essentially correct answer could fail.

\subsection{Guidance and task span}
\label{sec:guidance-span}

Our construction lets us assemble variants of the benchmark.
We vary methodological guidance through the level of detail in single-task
prompts, and task span by combining single tasks into workflows.
Crossing them gives the four conditions SF, SR, WF and WR introduced in
Section~\ref{sec:introduction}.

\paragraph{Full and reduced guidance.}
The full-guidance prompt states the scientific target, the inputs, the
method and the required output.
The reduced-guidance prompt removes method guidance that the grader does not
check (Fig.~\ref{fig:task-instance}, shaded) and keeps everything else, so
both prompts share one grader.
In 42 tasks the full prompt contains no such guidance, and the two prompts
are identical.
Of the other 52 tasks, 34 lose explicit method or implementation guidance
and 18 lose explanations or hints.

\paragraph{Single tasks and workflows.}
Because the selected steps keep their scientific dependencies, we can chain
single tasks from one study into a workflow that takes the agent from the
initial inputs to the final result (Fig.~\ref{fig:task-instance}, right).
We build eight workflows from eight studies (Table~\ref{tab:workflows}).
Seven have three stages, and one has four.
Unlike single tasks, which cover one step each, a workflow requires the
agent to run every calculation itself, including simulations
in seven of the eight workflows, within a five-hour limit on one node.
\rev{Our precomputed outputs are not given to the agent in a workflow.
The role of our precomputation is to ensure that the whole chain is
reproducible and to provide the reference for grading.}
Each workflow has full- and reduced-guidance prompts, and fixed rules grade
its final quantities against our reference.

\subsection{Composition}
\label{sec:composition}

\begin{figure}[t]
\centering
\includegraphics[width=\textwidth]{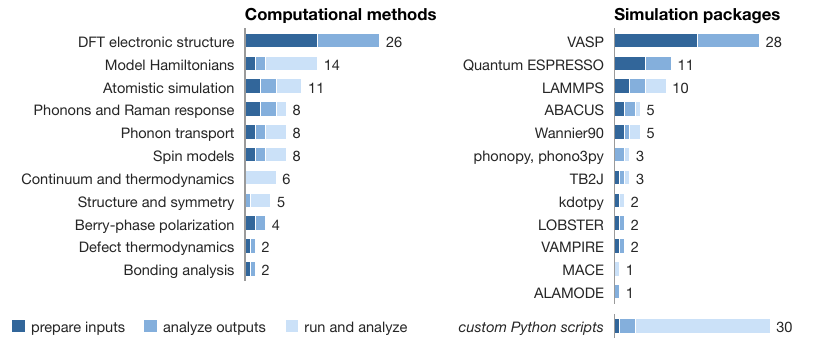}
\caption{Composition of the 94 tasks. Left: tasks by computational method.
Right: tasks by the scientific code they involve.
Shading gives the kind of task, as in Fig.~\ref{fig:benchmark-overview}.}
\label{fig:task-composition}
\end{figure}

CompMat-Bench covers different journals, computational methods and
scientific codes.
The 15 source studies appeared in Physical Review B (11), Nano Letters (2)
and npj Computational Materials (2).
Their 94 tasks span 11 method families and 12 scientific codes
(Fig.~\ref{fig:task-composition}).
VASP \citep{kresse1996vaspprb} (28 tasks), Quantum ESPRESSO
\citep{giannozzi2009quantumespresso} (11) and LAMMPS \citep{thompson2022lammps}
(10) are the most common codes, and in 30 tasks the agent writes the calculation in Python without
a named code.
The tasks comprise 30 input-preparation, 29 output-analysis and 35
run-and-analyze tasks, and no study contributes more than 9.

\section{Evaluation Setup}
\label{sec:evaluation-setup}

\paragraph{Agents.}
We evaluate \rev{ten} agents built on \rev{nine} LLMs.
DeepSeek V4.1 Flash \citep{deepseekai2026deepseekv41flash}, Gemini 3.8 Flash
\citep{googledeepmind2026gemini38flash}, GPT-5.6 Sol
\citep{openai2026gpt56systemcard}\rev{, GPT-6.1 Sol
\citep{openai2026gpt61soladdendum}, GPT-6.2 Astra
\citep{openai2026gpt62systemcard}, Gemini 4 Argon
\citep{googledeepmind2026gemini4argon}, Gemini 4 Carbon
\citep{googledeepmind2026gemini4carbon}, Qwen3.8-Max
\citep{alibabacloud2026qwen38max0902} and GLM-5.3 \citep{zai2026glm53}}
each run in the MatClaw harness
\citep{matclaw}, and GPT-5.6 Sol also runs in OpenAI's Codex CLI
\citep{openai2025codexcli}.
In MatClaw, all models use high reasoning effort and, on single tasks,
at most 500 agent steps (100 for DeepSeek V4.1 Flash), with an 800k-token
context for DeepSeek V4.1 Flash and 256k tokens for \rev{the other models}.
Codex CLI also uses high reasoning effort and a 256k-token context, with no
step limit (Table~\ref{tab:agent-config}).
\rev{Sections~\ref{sec:results-single} and~\ref{sec:results-failures} analyze
DeepSeek V4.1 Flash, Gemini 3.8 Flash and GPT-5.6 Sol;
Appendix~\ref{app:added-models} gives the results of all nine models.}

\paragraph{Run conditions.}
The agent works in an isolated container, with a new session for every
run and no internet access except the model API.
It gets the task files and offline user manuals for the simulation codes the
task uses, and cannot see reference answers or graders.
For a single task, it has 4 CPU cores, 16~GiB of memory (6~GiB for 85 of the
94 DeepSeek V4.1 Flash runs under SF) and one hour, and expensive simulations
are precomputed.
For a workflow, the agent runs all calculations itself on one compute node with
two 64-core AMD EPYC 7763 processors and 239~GiB of memory, and must finish
within five hours and 5,000 steps.
\rev{Appendix~\ref{app:completion-times} gives the completion times of the
agents.}

\paragraph{Evaluations and scoring.}
We run the following evaluations:
(1) SF: each of the \rev{ten} agents runs all 94 tasks once, and \rev{DeepSeek V4.1
Flash, Gemini 3.8 Flash and GPT-5.6 Sol in MatClaw run} a selected subset of
20 tasks five more times to measure the reliability of \rev{their}
performance;
(2) SR: each MatClaw agent \rev{except GPT-6.2 Astra and Gemini 4 Carbon} runs
once the 52 tasks whose reduced prompt removes
method guidance or explanations;
(3) WF and WR: \rev{the same seven agents run} each of the eight workflows under both
guidance levels in three rounds, 24 runs per agent and condition.
We report the pass rate, the fraction of runs whose submission meets every
check of the task's grading rules (Section~\ref{sec:studies-to-tasks}); for a
workflow, the checks apply to its final result.

\section{Results and Analysis}
\label{sec:results}

\subsection{All agents pass most single tasks under full guidance}
\label{sec:results-single}

\rev{The four agents of Table~\ref{tab:system-results}} pass most single tasks
under full guidance, with pass rates
ranging from 66.0 to 91.5\% (Table~\ref{tab:system-results} and
Fig.~\ref{fig:complete-results})\rev{, and the other six models pass 62.8 to
97.9\% (Table~\ref{tab:added-models})}.
GPT-5.6 Sol passes 90.4\% in MatClaw and 91.5\% in Codex CLI, so the two
harnesses give nearly the same result for the same model.
Input preparation is the hardest kind of task for \rev{each of these four
agents}, with pass
rates of 36.7 to 76.7\%, whereas the three stronger agents pass 96.6 to
100\% of the output-analysis and run-and-analyze tasks.
At standard vendor prices, one run over the 94 tasks costs 3.86~USD for
DeepSeek V4.1 Flash, 27.00~USD for GPT-5.6 Sol and 106.43~USD for Gemini 3.8
Flash.
GPT-5.6 Sol thus matches Gemini 3.8 Flash in pass rate at a quarter of the
cost, despite a higher price per token.
The high cost of Gemini 3.8 Flash comes from its long runs: it takes a median
of 56 steps per task, against 4 for GPT-5.6 Sol, and uses 48 times as many
input tokens.

To test whether one run per task is enough, we also run repeated tests with
\rev{DeepSeek V4.1 Flash, Gemini 3.8 Flash and GPT-5.6 Sol in MatClaw} on a
selected subset of tasks (five runs on 20 tasks;
Table~\ref{tab:repeatability}).
The ranking of Table~\ref{tab:system-results} is robust to repeated runs:
DeepSeek V4.1 Flash, Gemini 3.8 Flash and GPT-5.6 Sol pass 62, 88 and 93 of
the 100 runs, in the same order as in their single runs.
The failures of Gemini 3.8 Flash and GPT-5.6 Sol are mostly tied to specific
tasks rather than to chance between runs, since only 2 to 3 of the 20 tasks
give mixed results, passing some but not all five runs, whereas most failures
of DeepSeek V4.1 Flash vary between runs, with mixed results on 11 tasks.
\rev{The pass counts show no clear dependence on the publication date of the
source study (Appendix~\ref{app:publication-date}).}

\begin{table}[t]
\centering
\caption{Pass rates of four agents on the 94 single tasks with full guidance,
one run per task. API cost is the cost of the 94 runs at each vendor's
standard list price; the Codex CLI run was billed by subscription and has no
recorded cost.}
\label{tab:system-results}
\small
\setlength{\tabcolsep}{5pt}
\begin{tabular}{@{}llccccr@{}}
\toprule
 & & \multicolumn{4}{c}{Pass rate (\%)} & \\
\cmidrule(lr){3-6}
 & & Prepare & Analyze & Run and & All & API cost \\
Model & Harness & inputs & outputs & analyze & tasks & (USD) \\
 & & \footnotesize $n=30$ & \footnotesize $n=29$ & \footnotesize $n=35$ & \footnotesize $n=94$ & \\
\midrule
DeepSeek V4.1 Flash & MatClaw & 36.7 & 79.3 & 80.0 & 66.0 (62/94) & 3.86 \\
Gemini 3.8 Flash & MatClaw & 70.0 & 96.6 & 100.0 & 89.4 (84/94) & 106.43 \\
GPT-5.6 Sol & MatClaw & 76.7 & 96.6 & 97.1 & 90.4 (85/94) & 27.00 \\
GPT-5.6 Sol & Codex CLI & 76.7 & 100.0 & 97.1 & 91.5 (86/94) & -- \\
\bottomrule
\end{tabular}

\end{table}

\begin{table}[t]
\centering
\caption{Repeatability of three agents on a selected subset of 20 single
tasks with full guidance, five runs per task. The pass rate is over the 100
runs, and the remaining columns count the tasks by how many of the five runs
passed.}
\label{tab:repeatability}
\small
\begin{tabular}{@{}lccccccc@{}}
\toprule
 & & \multicolumn{6}{c}{Tasks by passing runs} \\
\cmidrule(l){3-8}
Model & Pass rate (\%) & 5/5 & 4/5 & 3/5 & 2/5 & 1/5 & 0/5 \\
\midrule
DeepSeek V4.1 Flash & 62.0 & 7 & 3 & 2 & 3 & 3 & 2 \\
Gemini 3.8 Flash & 88.0 & 16 & 1 & 1 & 0 & 1 & 1 \\
GPT-5.6 Sol & 93.0 & 17 & 2 & 0 & 0 & 0 & 1 \\
\bottomrule
\end{tabular}

\end{table}

\subsection{\texorpdfstring{\rev{Workflows with reduced guidance are the hardest condition, and single tasks predict the workflow ranking}}{Workflows with reduced guidance are the hardest condition, and single tasks predict the workflow ranking}}
\label{sec:results-conditions}

\begin{figure}[t]
\centering
\includegraphics[width=\textwidth]{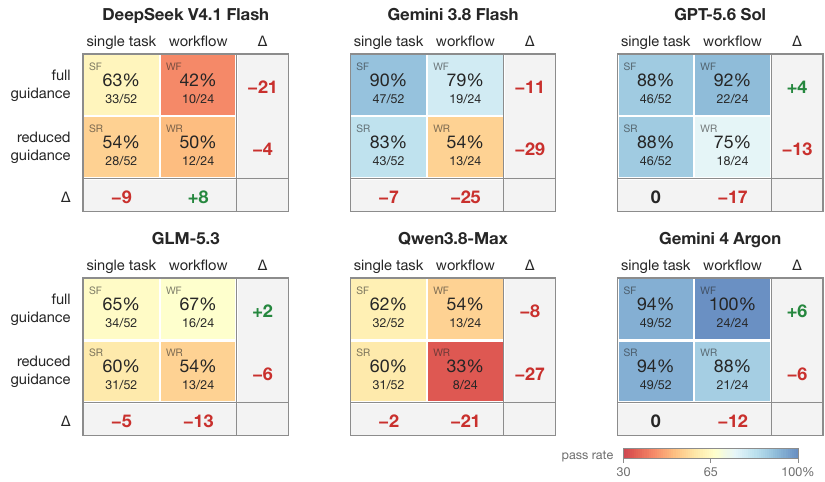}
\caption{Pass rates of \rev{six} agents under the four evaluation conditions of
Section~\ref{sec:guidance-span}.
Each cell gives the pass rate and the count behind it.
The single-task column covers the 52 tasks whose reduced prompt differs from
the full prompt, and the workflow column covers the eight workflows with three
runs each.
The $\Delta$ cells give differences in percentage points: reduced minus full
guidance (bottom row) and workflow minus single task (right column, different
task sets).
\rev{Results for more models are in Appendix~\ref{app:added-models}
(Table~\ref{tab:added-models}).}}
\label{fig:condition-grid}
\end{figure}

Fig.~\ref{fig:condition-grid} compares the pass rates of \rev{six} agents based
on different LLMs under four conditions: SF, SR, WF and WR.
In WF and WR, each of the eight workflows is run three times (24 runs per
condition).
In SF and SR, we cover the 52 of the 94 tasks that have a reduced prompt, so
that the two guidance levels are compared on the same tasks.
From Fig.~\ref{fig:condition-grid}, we find two conclusions common to all
agents.
(1) Longer task span and reduced guidance do not raise the pass rate beyond
noise, and mostly lower it.
Compared with SF, WF lowers the pass rates of DeepSeek V4.1 Flash and Gemini
3.8 Flash by 21 and 11 points, and SR lowers them by 9 and 7 points, whereas
GPT-5.6 Sol\rev{, GLM-5.3, Qwen3.8-Max and Gemini 4 Argon stay within 8 points}
of SF under both.
(2) WR is the hardest condition, with pass rates of 50, 54 and 75\% for
DeepSeek V4.1 Flash, Gemini 3.8 Flash and GPT-5.6 Sol\rev{, and of 54, 33 and
88\% for GLM-5.3, Qwen3.8-Max and Gemini 4 Argon} (for DeepSeek V4.1
Flash, the WR pass rate exceeds the WF pass rate by two runs, within
statistical noise).

\rev{DeepSeek V4.1 Flash, Gemini 3.8 Flash and GPT-5.6 Sol all} perform worse
under WR than under SF, but for different
reasons: the weaker agents lose on longer workflows, and the stronger agent
only on the combination with reduced guidance.
For DeepSeek V4.1 Flash, longer workflows alone reduce its performance
markedly: the pass rate decreases by 21 points from SF to WF, both under full
guidance.
We speculate that this agent has difficulty maintaining consistency over long
trajectories (11 of its 26 failed workflow runs reach the five-hour limit).
In contrast, for GPT-5.6 Sol, the pass rates under WF and SR are nearly
identical to that under SF, and the pass rate decreases, by 13 points, only
when a long task span is combined with reduced guidance, which requires the
agent to plan and carry out the multi-step calculation itself.
\rev{The additional failures cannot be attributed to either factor alone, and
they include both wrong physics and wrong implementation
(Fig.~\ref{fig:failure-categories}).}
\rev{Gemini 4 Argon shows the same pattern: it passes all 24 runs under WF and
21 under WR.}
Gemini 3.8 Flash falls between these two cases: longer workflows and reduced
guidance lower its pass rate by 11 and 7 points, respectively.
Their combination lowers it by 36 points, far more than either alone, and its
WR pass rate (54\%) is close to that of DeepSeek V4.1 Flash (50\%).

\rev{Seven models have results under all four conditions
(Table~\ref{tab:added-models}), which allows us to test how well the single
tasks predict the workflows.
We compare the two at the same level of guidance: SF with WF, and SR with WR.
With full guidance, the ranking in SF is GPT-6.1 Sol $>$ Gemini 4 Argon $>$
GPT-5.6 Sol $>$ Gemini 3.8 Flash $>$ Qwen3.8-Max $>$ GLM-5.3 $=$ DeepSeek V4.1
Flash, and only Qwen3.8-Max and GLM-5.3 exchange places in WF (Spearman rank
correlation 0.93).
With reduced guidance, the rank correlation between SR and WR is 0.92.
The single tasks thus predict how the models rank on the workflows, at about
2\% of the compute cost: for a typical model, the 94 single tasks take about
75 core-hours in total, and the 24 workflow runs about 3,300 core-hours in
total.}

\subsection{Workflow failures are mostly scientific, and agents share some interface errors}
\label{sec:results-failures}

\begin{figure}[t]
\centering
\includegraphics[width=\textwidth]{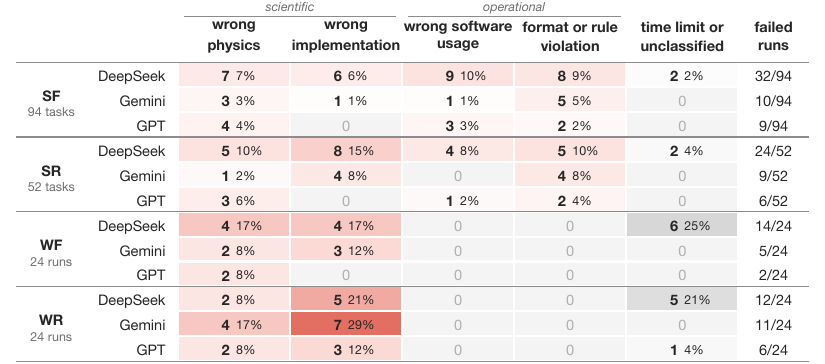}
\caption{Failure analysis: failed runs of three agents by cause under the
four evaluation conditions of Section~\ref{sec:guidance-span}.
DeepSeek, Gemini and GPT denote DeepSeek V4.1 Flash, Gemini 3.8 Flash and
GPT-5.6 Sol.
Each cell gives the number of failed runs and their share of all runs in that
row.}
\label{fig:failure-categories}
\end{figure}

We classify each failed run \rev{of DeepSeek V4.1 Flash, Gemini 3.8 Flash and
GPT-5.6 Sol} into three groups (Fig.~\ref{fig:failure-categories}):
scientific errors, either a wrong physical choice (for example, a missing
dipole correction for a polar slab) or a correct choice implemented wrongly
(for example, a sign error in a Bloch phase); operational errors, either
wrong software usage (for example, an input the code cannot parse) or a
format or rule violation (for example, a missing answer field); and runs that
reach the time limit or remain unclassified.
In workflows, 38 of the 50 failed runs are scientific errors, and 11 of the
other 12 are time limits of DeepSeek V4.1 Flash.
In single tasks, scientific and operational errors are about equal, 42 and 44
of 90.
All 18 software-usage errors occur in input-preparation tasks, where the
agent writes the inputs without running the software and therefore receives
no error message, whereas in a workflow the agent runs each code and can
correct an input that fails.
\rev{Table~\ref{tab:added-failures} gives the same classification for the other
six models.}

In-depth analysis suggests that agents fail to diagnose their errors because
they do not recognize anomalies that an experienced researcher would notice.
In the CrN workflow of Fig.~\ref{fig:task-instance}, 9 of the 15 failing
runs use inappropriate settings for the Wannier construction.
Their Wannier spreads reach 105 to 215~\AA$^2$, four to eight times the 25
to 28~\AA$^2$ of the passing runs, or the disentanglement does not converge,
and the exchange constants computed from them deviate by up to 116\%.
In each of these nine runs, the anomaly is printed in the agent's own
Wannier90 output, yet the agent continues to the exchange calculation without
correcting the settings.
We speculate that such diagnostic knowledge is implicit: it is rarely
documented in the literature and is instead acquired through research
practice, so it is poorly represented in the training data of language
models.

Some software interfaces are difficult for \rev{all four agents of
Table~\ref{tab:system-results}}.
For example, one task asks for the LAMMPS inputs of a WS$_2$/WSe$_2$
moir\'e bilayer with six atom types, and the line that sets the interaction
between the two layers must give an element name for each of the six types.
The offline LAMMPS manual is available to the agents, but it states no rule
for the number of names and its example uses two atom types, and \rev{each of
them} keeps this form, with two element names instead of six.
In the repeated runs, all 15 trials make the same mistake.
Because the task asks for inputs without running LAMMPS, no error message
reveals the mistake.
Agents also fail when a calculation needs a setting beyond the code
defaults, such as the van der Waals correction for a bilayer, which all four
agents omit.
Like the diagnostic knowledge above, the knowledge these interfaces require
is implicit: the documentation available to the agents does not state it, so
agents based on different LLMs fail in the same way.

\section{Limitations}
\label{sec:limitations}

CompMat-Bench has the following limitations: the strongest current agents
pass most single tasks under full guidance \rev{(up to 92 of 94,
Table~\ref{tab:added-models})}, which leaves little room to
distinguish stronger models, and the conditions that still separate them,
workflows and reduced guidance, have fewer runs; fixed-rule grading can
wrongly reject a correct answer, and although our failure analysis found and
fixed several such cases, others may remain; and the tasks do not cover all
computational materials research, for example studies that propose new ideas
or concepts, which are excluded at selection.
We plan to add tasks to the workflow and reduced-guidance conditions in later
versions.

\section{Conclusion}
\label{sec:conclusion}

We introduce CompMat-Bench, a benchmark of 94 tasks for evaluating AI agents
on recently published computational materials research: agents are assessed
on preparing the inputs and analyzing the outputs of expensive simulations,
which we reproduced in advance, rather than on running them during
evaluation.
With full methodological guidance, agents based on \rev{nine} LLMs pass
\rev{62.8--97.9\%} of the single tasks.
Varying the task span and the level of methodological guidance shows
that both factors lower the pass rates of the weaker agents, whereas the
\rev{stronger agents fall} only when a long workflow is combined with reduced
guidance.
\rev{Scientific errors are the most frequent cause of failure}, and in-depth
analysis reveals common
failure scenarios: agents do not recognize anomalies in their own results
that an experienced researcher would notice, and agents based on different
LLMs lack the same implicit knowledge of software interfaces.
CompMat-Bench provides a basis for comparing agents on these steps of real
materials research and for analyzing their failure modes.

\section*{Reproducibility Statement}
The tasks and workflows, with their prompts, input files, graders, reference
answers and offline reference library, are released at
\url{https://github.com/cz2014/CompMat-Bench} together with the run records
of every agent attempt reported in this paper.
Appendices~\ref{app:studies} to~\ref{app:supplementary} list the source
studies and tasks, the agent configuration and software versions, the
workflows and complete results, the prompts, and the two supplementary
workflows of the released package.
The intermediate files of our reproductions, about 1~TB, are not released.

\section*{AI Use Statement}
In this work, we used generative AI tools for decomposing the source
studies, \rev{reproducing their calculations,} drafting task prompts and
graders, and failure analysis.
We have not used generative AI tools for grading or for the final
scientific decisions, including the acceptance of each task, its reference
answer and the conclusions drawn; other tasks with required disclosure are
not applicable to this work.
Additionally, we used generative AI tools for drafting and editing the
manuscript.
We have reviewed all AI-assisted work and take responsibility for its final
content, including text, claims or artifacts produced with the aid of
generative AI.

\section*{Acknowledgements}
This work was supported by the U.S. AFOSR grant FA9550-24-1-0093.

\bibliographystyle{iclr2027_conference}
\bibliography{introduction_refs,related_work_refs,construction_refs,evaluation_refs}

\clearpage
\appendix
\raggedbottom
\counterwithin*{table}{section}
\renewcommand{\thetable}{\thesection\arabic{table}}
\counterwithin*{figure}{section}
\renewcommand{\thefigure}{\thesection\arabic{figure}}

\section{Source studies and task list}
\label{app:studies}

Table~\ref{tab:studies} lists the 15 source studies of
Section~\ref{sec:studies-to-tasks}, and Table~\ref{tab:tasks} lists the 94
single tasks derived from them.
The study numbers of Table~\ref{tab:studies} identify the source study of
each task in Table~\ref{tab:tasks}.
Each task is identified by the publication month of its source study and a
sequence number within that month, followed by a short name.
For example, single task 013 of Fig.~\ref{fig:task-instance} is 202603\_013.

\begingroup
\noindent
\begin{minipage}{\textwidth}
\captionof{table}{The 15 source studies.
Month is the publication month.
Methods lists the method families of Fig.~\ref{fig:task-composition} for
the tasks of each study, shortened to their first words.
Graph steps counts the research steps in the full graph of each study,
Selected counts the steps in its selected sub-graph, and Tasks counts the single
tasks built from these steps (Section~\ref{sec:studies-to-tasks}).}
\label{tab:studies}
\small
\setlength{\tabcolsep}{4pt}
\begin{tabularx}{\textwidth}{@{}rllc>{\raggedright\arraybackslash}Xrrr@{}}
\toprule
No. & Study & Journal & Month & Methods & \multicolumn{1}{c}{\begin{tabular}[b]{@{}c@{}}Graph\\steps\end{tabular}} & Selected & Tasks \\
\midrule
1 & \citep{li2025undulation} & Phys. Rev. B & 2025-09 & DFT, Structure & 10 & 4 & 7 \\
2 & \citep{jiang2026altermagnetism} & Nano Lett. & 2026-01 & Model Hamiltonians, DFT & 7 & 5 & 5 \\
3 & \citep{xie2026heusler} & Phys. Rev. B & 2026-01 & DFT, Structure, Spin models & 10 & 5 & 8 \\
4 & \citep{devaraj2026mnte} & Phys. Rev. B & 2026-02 & DFT, Model Hamiltonians, Structure & 12 & 5 & 7 \\
5 & \citep{hofer2026electrostatic} & Nano Lett. & 2026-02 & Model Hamiltonians & 6 & 3 & 4 \\
6 & \citep{parlak2026raman} & Phys. Rev. B & 2026-02 & Phonons, DFT & 10 & 3 & 6 \\
7 & \citep{takagi2026geometry} & Phys. Rev. B & 2026-02 & Model Hamiltonians & 13 & 3 & 4 \\
8 & \citep{wang2026moire} & Phys. Rev. B & 2026-02 & DFT, Atomistic, Continuum & 18 & 5 & 8 \\
9 & \citep{xu2026transition} & npj Comput. Mater. & 2026-02 & Atomistic, Spin models & 7 & 3 & 3 \\
10 & \citep{chak2026ferroelectricity} & npj Comput. Mater. & 2026-03 & Berry phase, Continuum & 7 & 5 & 7 \\
11 & \citep{kim2026magnons} & Phys. Rev. B & 2026-03 & Spin models, DFT & 7 & 5 & 8 \\
12 & \citep{li2026scattering} & Phys. Rev. B & 2026-03 & Phonon transport, Structure & 11 & 7 & 9 \\
13 & \citep{taper2026vacancies} & Phys. Rev. B & 2026-03 & Defects, Bonding, DFT, Structure & 6 & 3 & 6 \\
14 & \citep{wijekoon2026dislocations} & Phys. Rev. B & 2026-03 & Atomistic & 8 & 4 & 6 \\
15 & \citep{wu2026isotope} & Phys. Rev. B & 2026-03 & Phonons, Continuum & 9 & 5 & 6 \\
\midrule
\multicolumn{5}{@{}l}{Total} & 141 & 65 & 94 \\
\bottomrule
\end{tabularx}

\end{minipage}
\endgroup

\begingroup
\noindent
\begin{minipage}{\textwidth}
\captionof{table}{The 94 single tasks.
A dash in Codes marks a task with no named code, where the agent writes the
calculation itself in Python.
Checks counts the checks in the grading rules of the task.
QE: Quantum ESPRESSO; phonopy includes phono3py.}
\label{tab:tasks}
\end{minipage}
\small
{\setlength{\tabcolsep}{4pt}
\begin{longtable}{@{}l>{\footnotesize\raggedright\arraybackslash}p{140.64pt}rl>{\raggedright\arraybackslash}p{2.2cm}r@{}}
\toprule
Task & \multicolumn{1}{l}{Short name} & Study & Kind & Codes & Checks \\
\midrule
\endfirsthead
\toprule
Task & \multicolumn{1}{l}{Short name} & Study & Kind & Codes & Checks \\
\midrule
\endhead
\bottomrule
\endlastfoot
202509\_001 & hbn-sliding-sweep-inputs & 1 & input preparation & VASP & 14 \\
202509\_002 & hbn-sliding-potential-step & 1 & output analysis & VASP & 4 \\
202509\_003 & hbn-undulated-sliding-ratio & 1 & run and analyze & -- & 8 \\
202509\_004 & bn-nanotube-inputs & 1 & input preparation & VASP & 15 \\
202509\_005 & bn-nanotube-flexoelectric-constant & 1 & output analysis & VASP & 1 \\
202509\_006 & hbn-flexo-sliding-decomposition-inputs & 1 & input preparation & VASP & 14 \\
202509\_007 & hbn-undulated-step-decomposition & 1 & output analysis & VASP & 14 \\
202601\_001 & v2coal-heusler-structures & 3 & run and analyze & -- & 2 \\
202601\_002 & v2coal-relax-energy-inputs & 3 & input preparation & VASP & 14 \\
202601\_003 & v2coal-formation-energies-ordering & 3 & output analysis & VASP & 10 \\
202601\_004 & v2coal-l21-abacus-inputs & 3 & input preparation & ABACUS & 12 \\
202601\_005 & v2coal-l21-magnetic-moment & 3 & output analysis & ABACUS & 2 \\
202601\_006 & v2coal-xa-abacus-inputs & 3 & input preparation & ABACUS & 13 \\
202601\_007 & v2coal-xa-site-moments & 3 & output analysis & ABACUS & 3 \\
202601\_008 & v2coal-heisenberg-exchange & 3 & run and analyze & ABACUS, TB2J & 5 \\
202601\_009 & snte-110-surface-dirac-cones & 2 & run and analyze & -- & 1 \\
202601\_010 & ruo2-snte-interface-inputs & 2 & input preparation & VASP & 20 \\
202601\_011 & coupled-slab-surface-splittings & 2 & run and analyze & -- & 2 \\
202601\_012 & surface-cone-hall-contributions & 2 & run and analyze & -- & 2 \\
202601\_013 & slab-hall-interface-distribution & 2 & run and analyze & -- & 2 \\
202602\_001 & hgte-well-hartree-inputs & 5 & input preparation & kdotpy & 15 \\
202602\_002 & hgte-well-hartree-analysis & 5 & output analysis & -- & 10 \\
202602\_003 & hgte-well-surface-dispersion & 5 & run and analyze & kdotpy & 5 \\
202602\_004 & hgte-well-sdh-densities & 5 & run and analyze & -- & 10 \\
202602\_005 & gete-wannier-inputs & 7 & input preparation & QE, Wannier90 & 29 \\
202602\_006 & gete-fermi-levels & 7 & output analysis & Wannier90 & 5 \\
202602\_007 & gete-interband-jdos & 7 & run and analyze & Wannier90 & 8 \\
202602\_008 & gete-optical-conductivity & 7 & run and analyze & Wannier90 & 5 \\
202602\_009 & rucl2-dftu-bands-inputs & 6 & input preparation & VASP & 17 \\
202602\_010 & rucl2-valley-gaps & 6 & output analysis & VASP & 4 \\
202602\_011 & rucl2-phonon-forces-inputs & 6 & input preparation & VASP & 14 \\
202602\_012 & rucl2-a1p-phonon-mode & 6 & output analysis & VASP, phonopy & 5 \\
202602\_013 & rucl2-frozen-phonon-optics-inputs & 6 & input preparation & VASP & 17 \\
202602\_014 & rucl2-magneto-raman-tensor & 6 & output analysis & VASP & 6 \\
202602\_015 & ws2-wse2-moire-lammps-inputs & 8 & input preparation & LAMMPS & 17 \\
202602\_016 & ws2-wse2-moire-reconstruction-fields & 8 & output analysis & LAMMPS & 17 \\
202602\_017 & ws2-wse2-bilayer-soc-band-inputs & 8 & input preparation & VASP & 18 \\
202602\_018 & ws2-wse2-band-offsets-at-k & 8 & output analysis & VASP & 5 \\
202602\_019 & tmd-monolayer-strain-sweep-inputs & 8 & input preparation & VASP & 13 \\
202602\_020 & tmd-strain-deformation-potentials & 8 & output analysis & VASP & 6 \\
202602\_021 & moire-piezoelectric-bound-charge & 8 & run and analyze & LAMMPS & 7 \\
202602\_022 & moire-screened-piezo-potential & 8 & run and analyze & -- & 6 \\
202602\_023 & mnte-altermagnet-pipeline-inputs & 4 & input preparation & QE & 17 \\
202602\_024 & mnte-altermagnet-spin-splitting & 4 & output analysis & QE & 6 \\
202602\_025 & doped-mnte-space-group-classification & 4 & run and analyze & -- & 2 \\
202602\_026 & doped-mnte-supercell-band-inputs & 4 & input preparation & QE & 18 \\
202602\_027 & doped-mnte-momentum-spin-splitting & 4 & output analysis & QE & 16 \\
202602\_028 & tb-chain-spin-splitting & 4 & run and analyze & -- & 9 \\
202602\_029 & substituted-tb-chain-spectrum & 4 & run and analyze & -- & 13 \\
202602\_030 & spin-lattice-hessian-minimum-modes & 9 & run and analyze & -- & 9 \\
202602\_031 & ni-island-eam-saddle-points & 9 & run and analyze & -- & 12 \\
202602\_032 & ni-island-mlp-saddle-certification & 9 & run and analyze & MACE & 8 \\
202603\_001 & zno-mgo-phase-boundaries & 10 & run and analyze & -- & 6 \\
202603\_002 & zno-finite-strain-inputs & 10 & input preparation & QE & 18 \\
202603\_003 & znmgo-elastic-piezoelectric & 10 & output analysis & QE & 28 \\
202603\_004 & znmgo-berry-phase-inputs & 10 & input preparation & QE & 16 \\
202603\_005 & znmgo-spontaneous-polarization & 10 & output analysis & QE & 7 \\
202603\_006 & znmgo-landau-devonshire-model & 10 & run and analyze & -- & 6 \\
202603\_007 & znmgo-strain-composition-phase-map & 10 & run and analyze & -- & 6 \\
202603\_008 & afm-chain-magnon-spectrum & 11 & run and analyze & -- & 2 \\
202603\_009 & crn-ldau-structure-inputs & 11 & input preparation & VASP & 14 \\
202603\_010 & crn-relaxed-structure-exchange-paths & 11 & output analysis & VASP & 15 \\
202603\_011 & crn-wannier-exchange-inputs & 11 & input preparation & VASP, Wannier90, TB2J & 29 \\
202603\_012 & crn-nearest-neighbour-exchange & 11 & output analysis & TB2J & 12 \\
202603\_013 & crn-linear-spin-wave-spectrum & 11 & run and analyze & -- & 2 \\
202603\_014 & crn-monte-carlo-deck & 11 & input preparation & VAMPIRE & 23 \\
202603\_015 & crn-neel-temperature & 11 & output analysis & VAMPIRE & 2 \\
202603\_016 & si-ge-interface-cells & 12 & run and analyze & -- & 4 \\
202603\_017 & si-ge-empirical-phonons & 12 & run and analyze & LAMMPS, phonopy & 16 \\
202603\_018 & si-ge-phonon-force-inputs & 12 & input preparation & VASP & 8 \\
202603\_019 & si-ge-abinitio-phonons & 12 & output analysis & VASP, phonopy & 16 \\
202603\_020 & si-ge-mode-resolved-transmission & 12 & run and analyze & -- & 45 \\
202603\_021 & si-ge-diffuse-mismatch-transmission & 12 & run and analyze & -- & 4 \\
202603\_022 & si-ge-reflection-specularity & 12 & run and analyze & -- & 4 \\
202603\_023 & si-ge-phonon-mc-tables & 12 & input preparation & -- & 12 \\
202603\_024 & si-ge-slab-thermal-conductivity & 12 & output analysis & -- & 8 \\
202603\_025 & hzo-disordered-supercell-inputs & 13 & input preparation & VASP & 15 \\
202603\_026 & hzo-oxygen-environments & 13 & output analysis & -- & 17 \\
202603\_027 & hzo-oxygen-vacancy-inputs & 13 & input preparation & VASP & 14 \\
202603\_028 & hzo-vacancy-charge-thermodynamics & 13 & output analysis & VASP & 55 \\
202603\_029 & hzo-bonding-analysis-inputs & 13 & input preparation & VASP, LOBSTER & 17 \\
202603\_030 & hzo-oxygen-sublattice-bonding & 13 & output analysis & LOBSTER & 46 \\
202603\_031 & batio3-hysteresis-lammps-inputs & 14 & input preparation & LAMMPS & 35 \\
202603\_032 & batio3-polarization-hysteresis & 14 & output analysis & LAMMPS & 7 \\
202603\_033 & batio3-dislocation-dipole-inputs & 14 & input preparation & LAMMPS & 22 \\
202603\_034 & batio3-dislocation-dipole-relaxed & 14 & output analysis & LAMMPS & 16 \\
202603\_035 & batio3-dislocation-strain-maps & 14 & run and analyze & LAMMPS & 14 \\
202603\_036 & batio3-dislocation-core-polarization & 14 & run and analyze & LAMMPS & 36 \\
202603\_037 & bas-force-constant-inputs & 15 & input preparation & QE & 14 \\
202603\_038 & bas-harmonic-force-constants & 15 & output analysis & QE, ALAMODE & 21 \\
202603\_039 & bas-natural-isotope-dielectric & 15 & run and analyze & -- & 11 \\
202603\_040 & bas-isotope-dielectric-dependence & 15 & run and analyze & -- & 12 \\
202603\_041 & bas-surface-phonon-polariton-gap & 15 & run and analyze & -- & 11 \\
202603\_042 & bas-near-field-heat-flux & 15 & run and analyze & -- & 19 \\
\end{longtable}
}

\endgroup

\section{Agent configuration and run environment}
\label{app:setup}

Table~\ref{tab:agent-config} gives the settings of each agent, and
Table~\ref{tab:software} the software available to it in the container.

\begingroup
\noindent
\begin{minipage}{\textwidth}
\captionof{table}{Agent configuration.
Steps is the step limit per single task, Context the context window, Output
the maximum output tokens per call, and Effort the reasoning effort.
For Codex CLI, none means that its native tool loop has no step cap, and the
dash means its default output limit.}
\label{tab:agent-config}
\small
\setlength{\tabcolsep}{5pt}
\begin{tabularx}{\textwidth}{@{}>{\raggedright\arraybackslash}Xlrrrl@{}}
\toprule
Model & Harness & Steps & Context & Output & Effort \\
\midrule
DeepSeek V4.1 Flash & MatClaw & 100 & 800k & 131,072 & high \\
\addlinespace
Gemini 3.8 Flash & MatClaw & 500 & 256k & 65,536 & high \\
\addlinespace
GPT-5.6 Sol & MatClaw & 500 & 256k & 128,000 & high \\
\addlinespace
GPT-5.6 Sol & Codex CLI 0.154.0 & none & 256k & -- & high \\
\addlinespace
GPT-6.2 Astra & MatClaw & 500 & 256k & 128,000 & high \\
\addlinespace
GPT-6.1 Sol & MatClaw & 500 & 256k & 128,000 & high \\
\addlinespace
Gemini 4 Argon & MatClaw & 500 & 256k & 65,536 & high \\
\addlinespace
Gemini 4 Carbon & MatClaw & 500 & 256k & 65,536 & high \\
\addlinespace
Qwen3.8-Max & MatClaw & 500 & 256k & 131,072 & high \\
\addlinespace
GLM-5.3 & MatClaw & 500 & 256k & 131,072 & high \\
\bottomrule
\end{tabularx}

\end{minipage}
\endgroup

MatClaw runs each attempt as a code-first loop: one step is one model reply,
a JSON object whose code field the harness executes in a persistent Python
kernel, returning the printed output as the next observation.
From that code the agent can call file, search and shell tools; web and
job-submission tools are disabled.
A run ends when the agent calls \texttt{final\_answer}, reaches its step
limit, or reaches the time limit.
Codex CLI receives the same task text and runs its native tool loop with web
search disabled.

Each run sees only its workspace: the task files, the offline user manuals
and read-only software.
The graders and reference values are kept outside the container and are
applied to the submission after the run.
The container has no network route except to the model API.
The manuals come from one reference library, so every run of every agent
reads the same documentation.

\begingroup
\noindent
\begin{minipage}{\textwidth}
\captionof{table}{Software in the container.
Single tasks and Workflows state whether the agent can run each simulation
code: yes, no, or source when only a read-only source tree is staged.
Where it cannot, the agent has the code's manual.
The Python packages are shared by every run.}
\label{tab:software}
\small
\setlength{\tabcolsep}{5pt}
\begin{tabularx}{\textwidth}{@{}>{\raggedright\arraybackslash}Xlll@{}}
\toprule
Simulation code & Version & Single tasks & Workflows \\
\midrule
VASP & 6.1.0; 6.4.1 & no & yes \\
Quantum ESPRESSO & 7.5 & no & yes \\
ABACUS & 3.10.1 & no & yes \\
Wannier90 & 3.1.0 & no & yes \\
LAMMPS & 22 Jul 2025, update 4 & yes & yes \\
phono3py, with phonopy & 4.4.0 & yes & yes \\
MACE, with PyTorch & 0.3.16; 2.13.0 (CPU) & yes & yes \\
ALAMODE & 1.4.2 & yes & yes \\
TB2J & 0.9.12.26 & yes & yes \\
TB2J & 0.9.20 & source & yes \\
kdotpy & 1.4.1 & source & no \\
VAMPIRE & 7.0.0 & source & no \\
LOBSTER & -- & no & no \\
\midrule
Python package & Version & Python package & Version \\
\midrule
Python & 3.12.14 & spglib & 2.7.0 \\
numpy & 2.4.6 & matplotlib & 3.10.9 \\
scipy & 1.17.1 & pandas & 3.0.3 \\
ase & 3.27.0 & h5py & 3.16.0 \\
pymatgen & 2026.5.4 & sympy & 1.14.0 \\
\bottomrule
\end{tabularx}

\end{minipage}
\endgroup

\section{Workflows and complete results}
\label{app:results}

Table~\ref{tab:workflows} defines the eight workflows of
Section~\ref{sec:guidance-span}: the source study, the steps in order, and
the single task that tests the same step.

\begingroup
\noindent
\begin{minipage}{\textwidth}
\captionof{table}{The eight workflows.
Study numbers refer to Table~\ref{tab:studies} and task numbers to
Table~\ref{tab:tasks}; a range such as 202601\_004 to 007 lists the input
and analysis tasks of one step.
A dash marks a step that has no single task.}
\label{tab:workflows}
\end{minipage}
\small
{\setlength{\tabcolsep}{4pt}
\begin{longtable}{@{}lcc>{\raggedright\arraybackslash}p{222.925pt}l@{}}
\toprule
Workflow & Study & Step & Description & Single task \\
\midrule
\endfirsthead
\toprule
Workflow & Study & Step & Description & Single task \\
\midrule
\endhead
\bottomrule
\endlastfoot
C01 & 8 & a & Build and relax the WS$_2$/WSe$_2$ moir\'e & 202602\_015 to 016 \\*
 &  & b & Compute the piezoelectric bound charge & 202602\_021 \\*
 &  & c & Compute the screened electron potential & 202602\_022 \\
\addlinespace
C02 & 4 & a & Classify Se-pair substitutions by space group & 202602\_025 \\*
 &  & b & Build a tight-binding model per family & -- \\*
 &  & c & Compute the spin splittings & 202602\_028 to 029 \\
\addlinespace
C03 & 15 & a & Compute BAs force constants from DFT forces & 202603\_037 to 038 \\*
 &  & b & Compute the isotope-disordered dielectric function & 202603\_039 \\*
 &  & c & Compute gap modes and near-field transmission & 202603\_041 \\
\addlinespace
C04 & 12 & a & Relax the Si/Ge interface and compute force constants & -- \\*
 &  & b & Compute the phonon reflection amplitudes & 202603\_020 \\*
 &  & c & Compute the specularity of reflection & 202603\_022 \\
\addlinespace
C05 & 10 & a & Relax five ordered Zn$_{1-x}$Mg$_x$O cells & 202603\_002 to 003 \\*
 &  & b & Compute Berry-phase polarization paths & 202603\_004 to 005 \\*
 &  & c & Fit the Landau free-energy model & 202603\_006 \\
\addlinespace
C06 & 11 & a & Compute the magnetic ground state and Wannier functions & 202603\_011 \\*
 &  & b & Derive the four exchange constants & 202603\_012 \\*
 &  & c & Compute the spin-wave energies & 202603\_013 \\
\addlinespace
C07 & 3 & a & Build the two chemical orderings of V$_2$CoAl & 202601\_001 \\*
 &  & b & Relax both orderings with DFT & 202601\_002 to 003 \\*
 &  & c & Compute the magnetic states of both orderings & 202601\_004 to 007 \\*
 &  & d & Compute nearest-neighbour exchange couplings & 202601\_008 \\
\addlinespace
C08 & 9 & a & Find the Ni-island saddle point with EAM & 202602\_031 \\*
 &  & b & Refine the saddle with a machine-learned potential & -- \\*
 &  & c & Certify its curvature and connected minima & 202602\_032 \\
\end{longtable}}

\endgroup

\begin{figure}[htbp]
\centering
\includegraphics[width=\textwidth]{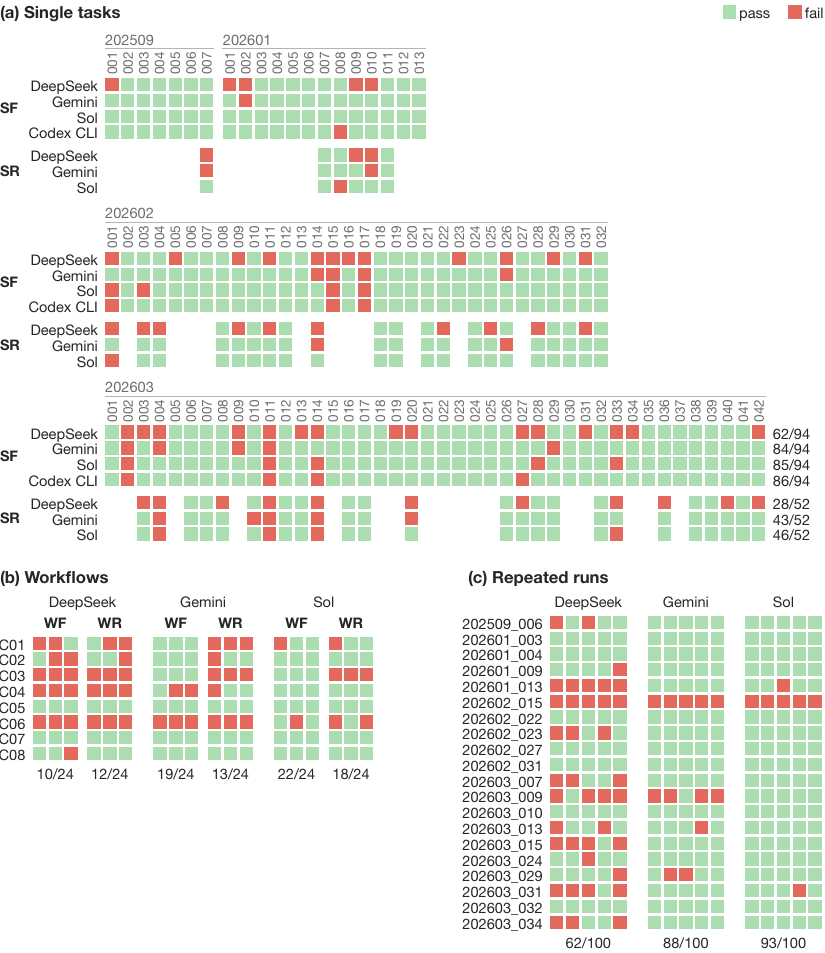}
\caption{Complete results.
Each square is one run.
DeepSeek and Gemini are DeepSeek V4.1 Flash and Gemini 3.8 Flash; Sol and
Codex CLI are GPT-5.6 Sol in the two harnesses of Table~\ref{tab:agent-config}.
(a) Columns are the 94 single tasks of Table~\ref{tab:tasks}, labelled by
task number.
SR rows are blank for the 42 tasks whose prompt has no reduced form.}
\label{fig:complete-results}
\end{figure}

Figure~\ref{fig:complete-results} gives every attempt behind
Table~\ref{tab:system-results}, Table~\ref{tab:repeatability} and
\rev{the top row of} Fig.~\ref{fig:condition-grid}: the 94 single tasks under SF and SR, the
eight workflows under WF and WR, and the 20 repeated tasks.

\clearpage
\section{Errors shared across agents}
\label{app:shared-errors}

This appendix lists the errors that several agents make in the same way on
the same task.
Section~\ref{sec:results-failures} shows one of them, a LAMMPS line that
all four agents write wrongly.
Of the 94 single tasks under full guidance, 14 are failed by two or more
agents.
Table~\ref{tab:shared-errors} lists six tasks on which the failing agents
write the same wrong input.
On five of these tasks the wrong input appears in every failing run; on the
sixth it appears in three of the four.

\begingroup
\noindent
\begin{minipage}{\textwidth}
\captionof{table}{Errors shared across agents on single tasks under full
guidance.
Task numbers refer to Table~\ref{tab:tasks}.
Agents names those that fail the task with the error shown; an agent not
named passes the task or fails it in another way.}
\label{tab:shared-errors}
\small
\setlength{\tabcolsep}{4pt}
\begin{tabular}{@{}>{\raggedright\arraybackslash}p{80pt}>{\raggedright\arraybackslash}p{48pt}>{\raggedright\arraybackslash}p{104pt}>{\raggedright\arraybackslash}p{72pt}>{\raggedright\arraybackslash}p{53pt}@{}}
\toprule
Task & Code & What the agents write & What the task needs & Agents \\
\midrule
202602\_015, WS$_2$/WSe$_2$ moir\'e bilayer, LAMMPS inputs & LAMMPS & Interlayer potential \texttt{kolmogorov/crespi/z} with two element names & One element name per atom type, six for this cell & all four \\
\addlinespace
202602\_017, WS$_2$/WSe$_2$ bilayer, band-structure inputs & VASP & Relaxation without a van der Waals correction (\texttt{IVDW} unset) & A dispersion correction for the van der Waals-bonded bilayer & all four \\
\addlinespace
202603\_011, CrN Wannier functions to exchange constants, inputs & VASP, Wannier90, TB2J & Spin-channel files named \texttt{wannier90.up} and \texttt{wannier90.dn} & The names VASP writes, \texttt{wannier90.1} and \texttt{wannier90.2} & DeepSeek, Sol, Codex CLI \\
\addlinespace
202603\_014, CrN Monte Carlo deck & VAMPIRE & No \texttt{unit-cell-category} in the material file, so only one Cr sublattice is built & Each sublattice assigned its own category & DeepSeek, Sol, Codex CLI \\
\addlinespace
202601\_002, V$_2$CoAl relaxation inputs & VASP & \texttt{LASPH} not set & \texttt{LASPH = .TRUE.} for 3d elements & DeepSeek, Gemini \\
\addlinespace
202603\_009, CrN DFT+U structure inputs & VASP & \texttt{LASPH} not set & \texttt{LASPH = .TRUE.} for 3d elements with DFT+U & DeepSeek, Gemini \\
\bottomrule
\end{tabular}

\end{minipage}
\endgroup

Section~\ref{sec:results-failures} names the cause: agents based on
different LLMs lack the same implicit knowledge of software interfaces.
Each row of Table~\ref{tab:shared-errors} is such a case.
The correct form is known from practice with the code: an example in the
manual does not generalise to more atom types, a default must be
overridden, or a tag must be set for a class of elements.
Given the same documentation, the agents converge on the same wrong form.

\clearpage
\section{Prompts}
\label{app:prompts}

\subsection{Decomposition prompt}
\label{app:prompts-decomposition}
The prompt below, given one source paper, produces the graph of research
steps of Section~\ref{sec:studies-to-tasks}.
\begin{Verbatim}[breaklines,fontsize=\scriptsize,frame=lines,framesep=4pt]
You are a research-paper expert specializing in methodological analysis, problem
    decomposition, and computational-dependency analysis of scientific studies.

**GOAL**
Produce, in a SINGLE response, a directed acyclic graph (DAG) of the paper's
research-problem decomposition. Work in two INTERNAL phases, then output once.

PHASE 1 -- Construct the research-problem tree (do not output yet):
- Fully comprehend the paper; understand its core research problems and experiments.
- Build a hierarchical research-problem tree that mirrors the authors' logic:
  * Root node -- the single, broadest research problem tackled by the paper.
  * Intermediate nodes -- progressively narrower sub-problems / objectives.
  * Leaf nodes -- fully specified experimental tasks (datasets, models, metrics,
    protocols) that each map to a MAIN-TEXT Figure or Table.
- Continue decomposing until every branch ends in such a leaf. There is no depth limit.

PHASE 2 -- Annotate the COMPLETE tree into a DAG (only after Phase 1 is finished, so the
full set of leaves is already known):
- For each leaf node, append: `id`, `cost_class`, `consumes`.
- For each intermediate node (root included), append: `id`.
- Because the whole tree already exists before you annotate, every `consumes` reference
  must point to a leaf you have already identified in Phase 1.

Output ONLY the final annotated JSON (the Phase 1 tree is internal scratch work).

---

### Phase 1 -- Reading & Extraction Rules
1. **Locate the root** in the title, abstract, introduction, or discussion.
2. **Recursively decompose** each problem by following explicit textual cues (headings,
    "first... second...", "to this end...", method overviews, figure/table captions,
    bullet lists, etc.).
3. **Identify leaves**: a node is a leaf *only if* it describes a concrete experiment and
    its `evidence.figure` or `evidence.table` field cites a **main-text** Figure or Table.
    A section reference alone is not sufficient. Supplementary material (SI, SM, Appendix,
    Supplementary Figs./Tables) may appear only inside a leaf's `protocol_or_setup` field.
4. **Not every main-text figure is a leaf.** A leaf produces a new physical quantity by
    running a calculation, a measurement, or its own equation on upstream output. The
    following are never leaves; fold each into the leaf whose output it reads, by adding
    the figure or table to that leaf's `evidence` and the quantity to its `metrics` and
    `conclusion`:
    (1) a figure or panel that only re-plots, visualizes, or takes summary statistics of
        another leaf's output;
    (2) a comparison of the paper's own results with literature or experimental values
        that the paper did not produce;
    (3) computing time, cost, or software-performance reports.
5. **Capture all layers** -- do **not** skip intermediate hypotheses, objectives, or
    analysis steps the paper explicitly discusses.
6. **Stay faithful** to the paper's wording for technical terms; paraphrase only for brevity.
7. **No outside invention** -- derive every node from the paper alone. If information is
    missing, mark the node with [uncertain].

### Phase 2 -- Annotation Rules
1. **Assign stable IDs.** Leaves receive `Lk` numbered top-to-bottom in tree order
    (`L1`, `L2`, ...). Intermediate nodes receive `Ik` (`I1`, `I2`, ...). The root node
    receives `id = "root"`.
2. **Tag `cost_class` for each leaf.** Choose `"heavy"` if the leaf describes a first-
    principles calculation (DFT, GW/BSE, AIMD, TDDFT), large-scale classical MD (~10^4
    atoms or longer than ~1 ns), Monte Carlo with non-trivial sampling, or ML-potential
    training -- anything taking more than ~1 hour on a single GPU/CPU node. Choose
    `"light"` otherwise: post-processing, analytical fits, geometry construction,
    structure mapping, visualization, or 1D ODE/PDE solving.
3. **Write `consumes` for each leaf.** List the IDs of upstream leaves whose outputs this
    leaf needs, e.g. `["L1", "L3"]`. Empty list `[]` is allowed for foundational leaves
    with no upstream. Dependencies may cross intermediate branches; do not invent
    dependencies that are not stated or implied in the paper.
4. **Stay acyclic and faithful.** The graph must be acyclic; a `consumes` reference that
    would create a cycle is the result of a misread and must be corrected. If a cost class
    or dependency is genuinely unclear, append `[uncertain]` to the relevant field.

Strictly output the annotated tree in a JSON format. Extraction fields come first; the
three annotation fields are appended within the same node object:
```
{ "paper": { "title": "", "authors": [], "venue": "", "year": "" },
  "problem_tree": {
    "node": "Root: broadest research problem tackled by the paper",
    "type": "root node",
    "description": "a detailed description of the research problem in this node",
    "evidence": "references back to the original paper",
    "id": "root",
    "children": [
      {"node": "Intermediate sub-problem / objective 1",
        "type": "depth-1 node",
        "description": "a detailed description of the research problem in this node",
        "evidence": "references back to the original paper",
        "id": "I1",
        "children": [
          {"type": "leaf node",
            "task": "Concrete experimental task (as phrased by paper)",
            "dataset": ["..."],
            "model_or_method": ["..."],
            "metrics": ["..."],
            "protocol_or_setup": "key settings/splits/hyperparams if stated",
            "evidence": {"figure": "Fig. X", "table": "Table Y", "section": "Sec. Z"},
            "conclusion": "explicit, detailed conclusions from this leaf's experiment",
            "status": "",
            "id": "L1",
            "cost_class": "heavy",
            "consumes": []
          }
        ]
      }
    ]
  }
}
```
\end{Verbatim}

\subsection{Time notice}
\label{app:prompts-time}
In a single-task run, MatClaw appends the notice below to the task and to
the observation returned after every step, with the remaining time updated;
the example is taken with 1,008 seconds left.
Codex CLI receives a similar notice.
\begin{Verbatim}[breaklines,fontsize=\scriptsize,frame=lines,framesep=4pt]
[Time budget] Total task wall-clock budget: 3600 seconds (60 minutes),
including API requests, reasoning, tool execution and waiting. Remaining at
this observation: 1008 seconds (16m 48s). Complete the required submission
files and call final_answer before the deadline.
\end{Verbatim}

\subsection{One task under reduced and full guidance}
\label{app:prompts-guidance}
The two prompts below are the reduced (SR) and full (SF) forms of task
202603\_020, the mode-resolved phonon transmission through a Si/Ge
interface.
The reduced prompt keeps the input file, the required accuracy, the
definitions of the reported quantities and the output format.
The full prompt adds the setting of the problem, the contents of the input
file, and the method: the atomistic Green's function approach, the criterion
for a propagating channel, and the conventions of the dynamical matrix and
of the normalised scattering matrix.
\begin{Verbatim}[breaklines,fontsize=\scriptsize,frame=lines,framesep=4pt,label=Reduced guidance (SR)]
# Task: mode-resolved phonon transmission through a Si/Ge interface from harmonic force constants

Compute the harmonic phonon transmission of a planar Si/Ge(100) interface
between a semi-infinite Si lead on the left and a semi-infinite Ge lead on
the right.

`device_force_constants.hdf5` contains the device's principal-layer
force-constant blocks and atomic masses. Its attributes specify units,
block ordering, coupling directions, layer period and transverse periodicity.
Use these blocks as given, without symmetrisation or sum-rule corrections.
No molecular-dynamics or electronic-structure calculation is required.
Python with numpy, scipy and h5py is available.

## Quantities and accuracy

Let $t$ be the flux-normalized Si-to-Ge transmission matrix, with incoming
propagating Si channels as rows and outgoing propagating Ge channels as columns.
For every incoming channel, transmission plus reflection must sum to one
with absolute error below $10^{-6}$.

At ordinary frequencies $\nu = 1.00, 2.00, 4.00, 5.05, 6.00, 8.00,
10.00, 11.50, 12.00$ THz, report:

- `n_Si`, `n_Ge`: the number of propagating channels in each lead travelling
  towards the interface;
- `Xi`: the sum of transmission probabilities over all incoming Si channels;
- `mean_T_Si`: the mean transmission probability over all incoming Si channels;
- `std_tau_Si`: the population standard deviation of the full spectrum of
  $t t^\dagger$ (all $n_{\mathrm{Si}}$ eigenvalues, including zeros).

## Output format

Write a single `answer.json` in the workspace root with exactly this layout:
one entry per requested frequency, in the order listed above, the frequency
echoed as given (counts are integers; the other three fields are dimensionless
floats):

```json
{
  "results": [
    {
      "frequency_THz": 1.00,
      "n_Si": <int>,
      "n_Ge": <int>,
      "Xi": <float>,
      "mean_T_Si": <float>,
      "std_tau_Si": <float>
    },
    ...
  ]
}
```

All values must be computed from the staged file; do not guess or hardcode.
\end{Verbatim}
\begin{Verbatim}[breaklines,fontsize=\scriptsize,frame=lines,framesep=4pt,label=Full guidance (SF)]
# Task: mode-resolved phonon transmission through a Si/Ge interface from harmonic force constants

A planar (100) interface between crystalline Si and crystalline Ge is studied
as a phonon scatterer. The interface sits in a device made of a semi-infinite
Si lead on the left, a semi-infinite Ge lead on the right and a central region
that contains the junction. All three parts are described by harmonic force
constants in a principal-layer basis. The question is how many propagating
phonon channels each lead carries towards the interface at a given frequency,
how much of the incident flux the interface transmits, and how unevenly the
transmission is distributed over the channels of the Si side.

## What you are given

- `device_force_constants.hdf5` -- the harmonic force-constant blocks of the
  device and the per-atom masses: for each lead, the on-site block of one
  principal layer and the coupling block to the next layer; for the central
  region, the on-site block of each of its slices, the couplings between
  consecutive slices, and the couplings of the first and last slice to their
  leads. Units, the atom-major ordering inside the blocks and the direction
  convention are recorded in the file's attributes. For both leads the coupling
  block connects a principal layer to the next one along $+x$, the direction
  from the Si lead towards the Ge lead, and in every coupling block the row
  block precedes the column block along $+x$. Both leads share the one layer
  period recorded in the file. The blocks already contain the full transverse
  periodic cell of the device, so no transverse wave-vector sampling is
  needed. Use the blocks as given, without further symmetrisation or sum-rule
  corrections.

No molecular-dynamics or electronic-structure calculation is required. The
environment provides Python with numpy, scipy and h5py. Solving one frequency
takes of the order of a minute on one core; the frequencies are independent.

## Method

Use the atomistic Green's function method for a lead-device-lead geometry with
the leads treated as semi-infinite, extended to a mode-resolved scattering
matrix: at each frequency, the propagating Bloch channels of each lead are
obtained from the lead's own blocks, the scattering matrix between the incoming
and outgoing propagating channels of the two leads is formed from the device
Green's function, and each entry is normalised by the channel group velocities
so that its squared modulus is a probability. A lead channel counts as
propagating when its Bloch factor across one layer period has unit modulus
(real wave vector); evanescent channels are removed before the scattering
matrix is formed. At the requested frequencies no channel lies at a band edge,
so the propagating and evanescent sets are well separated.

Conventions (fixed for this task):

- Frequencies are ordinary frequencies $\nu$ in THz; the dynamical problem is
  posed in the angular frequency $\omega = 2\pi\nu$.
- The dynamical blocks are the mass-weighted force constants
  $\Phi_{ij}/\sqrt{m_i m_j}$, converted to SI angular-frequency-squared units.
- The small positive imaginary part added to $\omega^2$ for the retarded
  Green's functions must be chosen small enough that, for every incoming
  channel, transmission plus reflection sums to one to better than $10^{-6}$.
- Let $t$ be the normalised transmission block of the scattering matrix from
  the incoming propagating channels of the Si lead (rows) to the outgoing
  propagating channels of the Ge lead (columns). For an incoming Si channel
  $l$, its transmission is $\Gamma_l = \sum_m |t_{lm}|^2$, and the **total
  transmission** is $\Xi(\nu) = \sum_l \Gamma_l$ over all incoming propagating
  channels of the Si lead.
- The **transmission eigenvalues** $\tau_k$ are the eigenvalues of the Hermitian
  matrix $t\,t^\dagger$; there are as many as incoming Si channels, and their
  sum equals $\Xi$. Their standard deviation is the population one (the
  variance normalised by the number of channels). Unlike the individual
  $\Gamma_l$, the $\tau_k$ do not depend on how a degenerate multiplet of
  channels is resolved into a basis.

## What to determine

At each of the nine frequencies $\nu \in \{1.00, 2.00, 4.00, 5.05, 6.00, 8.00,
10.00, 11.50, 12.00\}$ THz:

(1) $n_{\mathrm{Si}}$ and $n_{\mathrm{Ge}}$, the number of propagating channels
    of each lead that travel towards the interface;

(2) $\Xi$, the total transmission from the Si side;

(3) the mean of $\Gamma_l$ over the incoming Si channels, and the standard
    deviation of the transmission eigenvalues $\tau_k$.

## Output format

Write a single `answer.json` in the workspace root with exactly this layout:
one entry per requested frequency, in the order listed above, the frequency
echoed as given (counts are integers; the other three fields are dimensionless
floats):

```json
{
  "results": [
    {
      "frequency_THz": 1.00,
      "n_Si": <int>,
      "n_Ge": <int>,
      "Xi": <float>,
      "mean_T_Si": <float>,
      "std_tau_Si": <float>
    },
    ...
  ]
}
```

- `n_Si`, `n_Ge`: item (1).
- `Xi`: item (2).
- `mean_T_Si`, `std_tau_Si`: item (3).

All values must be computed from the staged file; do not guess or hardcode.
\end{Verbatim}

\section{\texorpdfstring{\rev{Results of all nine models}}{Results of all nine models}}
\label{app:added-models}
\begin{revblock}
Table~\ref{tab:added-models} gives the number of passes of the nine
models of Section~\ref{sec:evaluation-setup} under each condition.
GPT-6.2 Astra and Gemini 4 Carbon were run on the fully guided single tasks
only.

\begingroup
\noindent
\begin{minipage}{\textwidth}
\centering
\captionof{table}{Passes of nine models under the four conditions, ordered by
SF.
A dash marks a condition that was not run.
$^{*}$Gemini 4 Carbon did not finish 34 tasks within one hour; the value with
an asterisk gives these tasks 2.5 hours.}
\label{tab:added-models}
\small
\begin{tabular}{@{}lccccc@{}}
\toprule
Model & SF, 94 tasks & SF, 52 & SR, 52 & WF, 24 runs & WR, 24 runs \\
\midrule
GPT-6.2 Astra & 92 & 50 & -- & -- & -- \\
GPT-6.1 Sol & 91 & 50 & 49 & 24 & 21 \\
Gemini 4 Argon & 90 & 49 & 49 & 24 & 21 \\
GPT-5.6 Sol & 85 & 46 & 46 & 22 & 18 \\
Gemini 3.8 Flash & 84 & 47 & 43 & 19 & 13 \\
Qwen3.8-Max & 64 & 32 & 31 & 13 & 8 \\
GLM-5.3 & 62 & 34 & 31 & 16 & 13 \\
DeepSeek V4.1 Flash & 62 & 33 & 28 & 10 & 12 \\
Gemini 4 Carbon & 59 (86$^{*}$) & 31 (46$^{*}$) & -- & -- & -- \\
\bottomrule
\end{tabular}
\end{minipage}
\endgroup

\vspace{1em}
Table~\ref{tab:added-failures} gives the classification of
Section~\ref{sec:results-failures} for the other six models.
In workflows, 22 of the 52 failed runs are scientific errors, and 26 of the
other 30 are time limits, 23 of them of Qwen3.8-Max.
In single tasks, 71 of the 154 failed runs are time limits, 34 of them of
Gemini 4 Carbon, and scientific and operational errors are 27 and 55.

\begingroup
\noindent
\begin{minipage}{\textwidth}
\centering
\captionof{table}{Failure analysis of the other six models: failed runs
by cause under the evaluation conditions of Section~\ref{sec:guidance-span}
that were run for each model.
Each cell gives the number of failed runs and their share of all runs in that
row (94 runs under SF, 52 under SR, and 24 each under WF and WR).
$^{*}$Gemini 4 Carbon did not finish 34 tasks within one hour; the row with
an asterisk gives these tasks 2.5 hours.}
\label{tab:added-failures}
\footnotesize
\setlength{\tabcolsep}{3pt}
\begin{tabular}{@{}llcccccc@{}}
\toprule
Model & Cond. & Failed & \shortstack{Wrong\\physics} &
\shortstack{Wrong\\implementation} & \shortstack{Wrong software\\usage} &
\shortstack{Format or rule\\violation} & \shortstack{Time limit or\\unclassified} \\
\midrule
GPT-6.2 Astra & SF & 2 & 1 (1\%) & 0 & 0 & 1 (1\%) & 0 \\
\addlinespace
GPT-6.1 Sol & SF & 3 & 1 (1\%) & 0 & 0 & 2 (2\%) & 0 \\
 & SR & 3 & 1 (2\%) & 0 & 1 (2\%) & 1 (2\%) & 0 \\
 & WF & 0 & 0 & 0 & 0 & 0 & 0 \\
 & WR & 3 & 2 (8\%) & 1 (4\%) & 0 & 0 & 0 \\
\addlinespace
Gemini 4 Argon & SF & 4 & 1 (1\%) & 0 & 0 & 3 (3\%) & 0 \\
 & SR & 3 & 0 & 0 & 1 (2\%) & 2 (4\%) & 0 \\
 & WF & 0 & 0 & 0 & 0 & 0 & 0 \\
 & WR & 3 & 1 (4\%) & 2 (8\%) & 0 & 0 & 0 \\
\addlinespace
Qwen3.8-Max & SF & 30 & 3 (3\%) & 1 (1\%) & 6 (6\%) & 5 (5\%) & 15 (16\%) \\
 & SR & 21 & 1 (2\%) & 1 (2\%) & 3 (6\%) & 4 (8\%) & 12 (23\%) \\
 & WF & 11 & 0 & 1 (4\%) & 0 & 1 (4\%) & 9 (38\%) \\
 & WR & 16 & 1 (4\%) & 0 & 1 (4\%) & 0 & 14 (58\%) \\
\addlinespace
GLM-5.3 & SF & 32 & 6 (6\%) & 2 (2\%) & 10 (11\%) & 8 (9\%) & 6 (6\%) \\
 & SR & 21 & 3 (6\%) & 5 (10\%) & 4 (8\%) & 4 (8\%) & 5 (10\%) \\
 & WF & 8 & 4 (17\%) & 2 (8\%) & 0 & 0 & 2 (8\%) \\
 & WR & 11 & 2 (8\%) & 6 (25\%) & 2 (8\%) & 0 & 1 (4\%) \\
\addlinespace
Gemini 4 Carbon & SF & 35 & 1 (1\%) & 0 & 0 & 0 & 34 (36\%) \\
 & SF$^{*}$ & 8 & 1 (1\%) & 0 & 0 & 0 & 7 (7\%) \\
\bottomrule
\end{tabular}
\end{minipage}
\endgroup
\end{revblock}

\section{\texorpdfstring{\rev{Grading checks}}{Grading checks}}
\label{app:grading-checks}
\begin{revblock}
A check is one condition in the grading rules of a task, and a task passes
only when all of its checks pass (Section~\ref{sec:studies-to-tasks}).
Table~\ref{tab:grading-checks} counts the 1,191 checks of the 94 tasks by the
four kinds defined there.
Exact matches are 105 of them (9\%) and apply to answers with one correct
value; under the other three kinds (1,072 checks, 90\%), an answer that
differs from our reference can pass.

\begingroup
\noindent
\begin{minipage}{\textwidth}
\centering
\captionof{table}{Number of grading checks of each kind in the 94 tasks.}
\label{tab:grading-checks}
\small
\begin{tabularx}{\textwidth}{@{}l>{\raggedright\arraybackslash}Xr@{}}
\toprule
Kind of check & What is checked & Checks \\
\midrule
Range & a value within a numerical tolerance, or beyond a lower or upper bound & 696 \\
\addlinespace
Subset & the input file contains a decisive setting, or does not contain a forbidden one & 184 \\
\addlinespace
Physical equivalence & a setting from a set of equivalent ones, a structure equivalent to the reference by translation or symmetry, or values consistent within or between files & 192 \\
\addlinespace
Exact match & the answer of an analysis task equals the reference: an integer count, a categorical label, a sign or an index & 105 \\
\addlinespace
Others & a fixed ordering, protocol or syntax & 14 \\
\midrule
Total & & 1,191 \\
\bottomrule
\end{tabularx}
\end{minipage}
\endgroup
\end{revblock}

\section{\texorpdfstring{\rev{Pass counts by publication date}}{Pass counts by publication date}}
\label{app:publication-date}
\begin{revblock}
We test whether the pass counts depend on the publication date of the source
study.
We order the 94 tasks by the publication date of their study and split them
into an earlier and a later half of 47 tasks each.
Table~\ref{tab:publication-date} gives the SF passes of seven models in each
half.
The pass counts show no clear dependence on the publication date: for six
models the two halves differ by at most 2 tasks, and for Qwen3.8-Max by 6.

\begingroup
\noindent
\begin{minipage}{\textwidth}
\centering
\captionof{table}{SF passes on the earlier and the later half of the 94
tasks, ordered by the publication date of the source study.}
\label{tab:publication-date}
\small
\begin{tabular}{@{}lcc@{}}
\toprule
Model & Earlier half (47 tasks) & Later half (47 tasks) \\
\midrule
GPT-6.1 Sol & 45 & 46 \\
Gemini 4 Argon & 45 & 45 \\
GPT-5.6 Sol & 43 & 42 \\
Gemini 3.8 Flash & 43 & 41 \\
Qwen3.8-Max & 35 & 29 \\
GLM-5.3 & 30 & 32 \\
DeepSeek V4.1 Flash & 32 & 30 \\
\bottomrule
\end{tabular}
\end{minipage}
\endgroup
\end{revblock}

\section{\texorpdfstring{\rev{Completion times}}{Completion times}}
\label{app:completion-times}
\begin{revblock}
Table~\ref{tab:completion-times} gives the median wall-clock time of the
passing runs of seven models.

\begingroup
\noindent
\begin{minipage}{\textwidth}
\centering
\captionof{table}{Median wall-clock time of passing runs, in minutes.}
\label{tab:completion-times}
\small
\begin{tabular}{@{}lcc@{}}
\toprule
Model & SF (94 single tasks) & WF (24 workflow runs) \\
\midrule
GPT-6.1 Sol & 1.5 & 38 \\
Gemini 4 Argon & 18.1 & 63 \\
GPT-5.6 Sol & 3.4 & 25 \\
Gemini 3.8 Flash & 7.1 & 33 \\
Qwen3.8-Max & 14.9 & 195 \\
GLM-5.3 & 4.9 & 84 \\
DeepSeek V4.1 Flash & 2.9 & 190 \\
\bottomrule
\end{tabular}
\end{minipage}
\endgroup
\end{revblock}

\clearpage
\section{Two supplementary workflows}
\label{app:supplementary}

The released package contains two further workflows, C09 and C10, beyond the
eight of Table~\ref{tab:workflows}.
Both are built from our own research projects rather than from published
studies, so they have no source study in Table~\ref{tab:studies} and no
single-task counterparts in Table~\ref{tab:tasks}.
Each has the same form as the eight: a prompt under full and reduced
guidance, the input files, a grader, and a reference answer from our own
calculation.

C09 comes from our study of a layered ferroelectric with a machine-learned
interatomic potential \citep{matclaw}.
The agent relaxes the bulk structure, compresses it along the stacking
direction, and runs molecular dynamics at a series of temperatures to locate
the loss of polar order; it reports the transition temperature of the
unstrained crystal and its rate of change with compression.
The reference values come from our own molecular dynamics of finite
duration and are an empirical estimate of the transition temperature, not a
converged thermodynamic value.
C10 comes from our study of energy transfer along molecular chains.
The agent computes from first principles the reaction energy and the barrier
of the isomerization of a small molecule, fits a one-dimensional double-well
potential to these two quantities, and integrates a damped chain of such
units to obtain the speed of the conversion front at three coupling
strengths.
The chain is a model study driven by molecular parameters, and its reference
values follow deterministically from the fitted potential and the equations
of motion.

\end{document}